\documentclass[letterpaper]{article} 
\usepackage{aaai2026}  
\usepackage{times}  
\usepackage{helvet}  
\usepackage{courier}  
\usepackage[hyphens]{url}  
\usepackage{graphicx} 
\usepackage{natbib}  
\usepackage{caption} 
\usepackage{algorithm}
\usepackage{algorithmic}

\usepackage{newfloat}
\usepackage{listings}
\DeclareCaptionStyle{ruled}{labelfont=normalfont,labelsep=colon,strut=off} 
\floatstyle{ruled}
\newfloat{listing}{tb}{lst}{}
\floatname{listing}{Listing}
\title{HCF: Hierarchical Cascade Framework for Distributed Multi-Stage Image Compression}
\author {
    Junhao Cai\textsuperscript{\rm 1},
    Taegun An\textsuperscript{\rm 1},
    Chengjun Jin\textsuperscript{\rm 1},
    Sung Il Choi\textsuperscript{\rm 1},
    Juhyun Park\textsuperscript{\rm 1},
    Changhee Joo\textsuperscript{\rm 1\thanks{Corresponding Author}}
}
\affiliations {
    \textsuperscript{\rm 1}Korea University, Seoul, Republic of Korea\\
    \{junhochae,antaegun20,chengjunjin2001,sungchoi,juhyunpark,changhee\}@korea.ac.kr
}

\usepackage{multirow}
\usepackage{amsmath}
\usepackage{booktabs}

\begin{document}

\maketitle

\begin{abstract}
Distributed multi-stage image compression\textemdash where visual content traverses multiple processing nodes under varying quality requirements\textemdash poses challenges. Progressive methods enable bitstream truncation but underutilize available compute resources; successive compression repeats costly pixel-domain operations and suffers cumulative quality loss and inefficiency; fixed-parameter models lack post-encoding flexibility. In this work, we developed the Hierarchical Cascade Framework (HCF) that achieves high rate-distortion performance and better computational efficiency through direct latent-space transformations across network nodes in distributed multi-stage image compression systems. Under HCF, we introduced policy-driven quantization control to optimize rate–distortion trade-offs, and established the edge quantization principle through differential entropy analysis. The configuration based on this principle demonstrates up to 0.6~dB PSNR gains over other configurations. When comprehensively evaluated on the Kodak, CLIC, and CLIC2020-mobile datasets, HCF outperforms successive-compression methods by up to 5.56\,\% BD-Rate in PSNR on CLIC, while saving up to 97.8\,\% FLOPs, 96.5\,\% GPU memory, and 90.0\,\% execution time. It also outperforms state-of-the-art progressive compression methods by up to 12.64\,\% BD-Rate on Kodak and enables retraining-free cross-quality adaptation with 7.13--10.87\,\% BD-Rate reductions on CLIC2020-mobile.
\end{abstract}

\section{Introduction}
\label{sec:intro}
The rapid growth of digital media consumption has significantly increased demands on visual information transmission systems~\cite{GHOSH2025108011, 6G_Visual_Data}, positioning image compression as an essential technology to address bandwidth and storage constraints while maintaining acceptable quality levels~\cite{ibraheem2024comprehensive}. Modern transmission scenarios involve multi-stage processing where visual content traverses multiple processing nodes during transmission~\cite{10.1145/3742472}. As deployment scenarios become increasingly complex, it is imperative for compression systems to provide multiple quality levels, enabling quality scaling during transmission~\cite{kong2024progressive}.
This creates scenarios where compression operations are distributed across multiple processing stages with varying quality levels—a paradigm we refer to as \textit{distributed multi-stage image compression}.

Within the distributed multi-stage compression framework, existing approaches have emerged but face fundamental limitations. Progressive compression methods~\cite{Lee_2022_CVPR,2023_CVPR_jeon,lee2025deephq,Presta_2025_WACV} enable quality adaptation through bitstream truncation but constrain intermediate processing to passive operations with suboptimal rate-distortion trade-offs. Successive compression approaches~\cite{SIC} achieve adaptation through repeated operations but suffer from cumulative quality degradation and computational inefficiency due to redundant pixel-domain conversions. Meanwhile, fixed-parameter compression models~\cite{minnenbt18,cheng2020image,jiang2023mlicpp}, while achieving superior rate-distortion performance in centralized settings, provide single operating points without post-encoding flexibility required for distributed multi-stage scenarios.

The emergence of distributed computational resources~\cite{MobileEdgeComputing2025,6g_edge,TangComputingNetworkConvergence,INCML,INC2023} in modern networks creates new opportunities for intelligent quality transformation within distributed multi-stage compression, necessitating methods that can efficiently leverage these capabilities while avoiding the performance penalties of existing approaches. To address this need, we propose the Hierarchical Cascade Framework (HCF), which harnesses distributed computational resources for efficient multi-stage image compression. To the best of our knowledge, this is the first framework to effectively leverage distributed computational resources for multi-stage compression by enabling direct latent-space transformations across network nodes and implementing policy-driven quantization control. This approach simultaneously improves compression performance and reduces computational overhead while preventing cumulative degradation through optimal placement strategies.

The contributions of our work are as follows:
\begin{itemize}
    \item We propose HCF, a novel framework facilitating distributed multi-stage image compression through direct latent-space transformations across network nodes, avoiding pixel-domain recompression cycles and introducing policy-driven quantization control.
    
    \item We systematically analyze and identify limitations in existing distributed multi-stage compression approaches: progressive methods suffer from passive bitstream truncation without leveraging intermediate computational resources, while successive methods utilize computational resources but incur redundant pixel-domain processing cycles.
    
    \item We discover and establish the edge quantization optimality principle, providing theoretical insights into optimal quantization placement that consistently outperforms alternatives by up to 0.6 dB PSNR, validated through differential entropy analysis.
    
    \item We provide comprehensive evaluation across five compression architectures, achieving substantial improvements over existing state-of-the-art (SOTA) baselines: outperforming successive compression methods by up to 5.56\% BD-Rate in PSNR and progressive compression methods by up to 12.64\% BD-Rate, achieving 97.1--97.8\% FLOPs reduction with 94.8--96.5\% GPU memory savings and 77.9--90.0\% execution time reduction, and enabling effective retraining-free cross-quality adaptation with 7.13--10.87\% BD-Rate reductions.
\end{itemize}

\section{Related Work}
\label{sec:related_work}
Image compression has evolved from traditional standards~\cite{jpeg,jpeg2000,hevc,webp,bpg} to deep learning approaches that achieve superior rate-distortion performance. This paradigm shift was marked by key milestones: autoencoder architectures with neural transforms~\cite{balle2016end,Lossy_2017}, entropy modeling through hyperprior networks~\cite{ballemshj18}, scale-mean Gaussians~\cite{minnenbt18}, and mixture models~\cite{cheng2020image,zhu2022unified,fu2023}, alongside context-aware processing that exploited local spatial dependencies~\cite{minnenbt18,he2022,liu2023learned,liu2024,jiang2023mlicpp}, global correlations~\cite{guo2021causal,khoshkhahtinat2023multi}, and channel-wise relationships~\cite{feng2025linear,zeng2025mambaic}, as well as dictionary-based modeling~\cite{wu2025conditional,lu2025learned} that leverages external priors. However, these learned methods remain designed as single-stage, centralized systems that generate static bitstreams, limiting their optimization potential in scenarios requiring post-encoding quality adaptation.

To address quality adaptation requirements, progressive compression approaches have evolved from early foundations with recurrent neural networks for variable compression rates~\cite{toderici2015variable,Toderici} and spatial adaptive bit allocation and diffusion mechanisms~\cite{Johnston}. These advances enabled scalable coding architectures with layer-wise bit allocation~\cite{su}, and fine granular scalability through nested quantization~\cite{lu2021progressive}, trit-plane encoding~\cite{Lee_2022_CVPR,2023_CVPR_jeon}, spatial autoregressive modeling with codeword alignment~\cite{Tian_VCIP}, multirate progressive entropy modeling that unifies split-coded-then-merge models~\cite{Li_tcsvt}, progressive deep coding with layer-specific parameter learning~\cite{lee2025deephq} and element-wise progressive transmission via variance-aware masking~\cite{Presta_2025_WACV}.

Meanwhile, successive image compression pipelines, where the same codec is repeatedly applied across stages via encoding–decoding cycles, can in principle produce multiple quality levels, but have been shown to suffer from cumulative quality degradation and instability~\cite{SIC}. In parallel, the emergence of distributed computational resources in modern networks—including 6G Edge Networks~\cite{6g_edge}, Mobile Edge Computing~\cite{choi2025cooperative,MobileEdgeComputing2025}, and In-Network Machine Learning~\cite{INCML}—highlights a broader shift towards exploiting intermediate nodes as active compute and decision points. These capabilities provide new opportunities for compression optimization through intermediate computational processing~\cite{10.1145/3742472}. However, existing compression methods cannot effectively leverage such resources: progressive methods are constrained to suboptimal truncation operations with passive intermediate nodes, whereas successive approaches incur computational redundancy through repeated pixel-domain operations.

\section{Methodology}
\label{sec:method}

\subsection{Existing Frameworks and Limitations}
\label{subsec:problem_formulation}

\textbf{Single-Stage Framework (SSF).}
Image compression systems traditionally operate under SSF, which follows a sequential pipeline comprising five fundamental operations. Let $x$ denote the input image, $\tilde{y}$ the unquantized latent representation, $\hat{y}$ the quantized latent representation, $\mathcal{B}$ the bitstream, and $\hat{x}$ the reconstructed image. These operations proceed as
\begin{align}
    \tilde{y} &= g_{a}(x), && \text{analysis transform,} \label{eq:analysis}\\
    \hat{y}   &= Q(\tilde{y}), && \text{quantization,} \label{eq:quantization}\\
    \mathcal{B} &= E(\hat{y}), && \text{entropy encoding,} \label{eq:encoding}\\
    \hat{y}   &= D(\mathcal{B}), && \text{entropy decoding,} \label{eq:decoding}\\
    \hat{x}   &= g_s(\hat{y}), && \text{synthesis transform.} \label{eq:synthesis}
\end{align}

\noindent
Here, $g_a(\cdot)$, $Q(\cdot)$, $E(\cdot)$, $D(\cdot)$, and $g_s(\cdot)$ denote the analysis transform, quantization, entropy encoding, entropy decoding, and synthesis transform operators, respectively, all applied under a consistent quality configuration with rate–distortion parameter $\lambda$. While effective for centralized compression, SSF lacks post-encoding adaptability required for distributed multi-stage scenarios.

For distributed multi-stage compression, two primary approaches have emerged:

\noindent
\textbf{Progressive Compression Framework (PCF).} 
PCF generates scalable bitstreams enabling quality adaptation through selective decoding and bitstream truncation. This ``encode once, decode many" paradigm allows intermediate nodes to modify compression rates by discarding specific portions of the transmitted bitstream. However, this bitstream-centric approach restricts intermediate nodes to predetermined truncation patterns, without the flexibility to exploit node-specific computational resources for adaptive rate-distortion optimization in distributed multi-stage scenarios.

\begin{figure}[t]
    \centering
    \resizebox{\linewidth}{!}{%
        \includegraphics{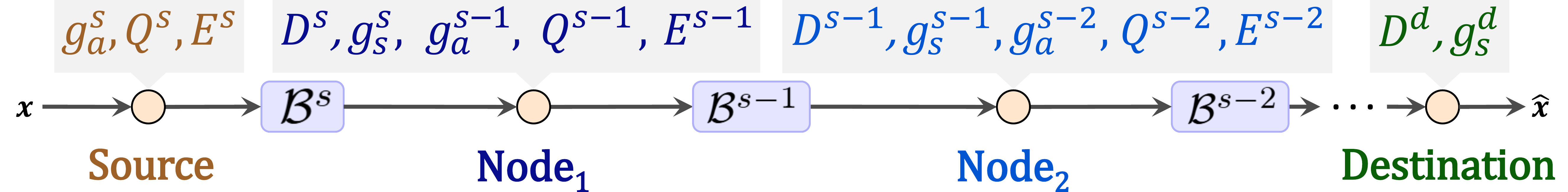}
    }
    \caption{The Distributed Recompression Framework (DRF) extends SIC to distributed compression across multiple nodes. It repeats the decompression-compression cycle at intermediate nodes. Specifically, each Node$_k$ compresses the image at quality level $s-k$ followed by the full decompression operations of quality level $s-k+1$.}
    \label{fig:Naive_distributed}
\end{figure}

\noindent
\textbf{Distributed Recompression Framework (DRF).} 
DRF extends Successive Image Compression (SIC)~\cite{SIC} to distributed multi-stage scenarios. Let $s$ denote the source quality level, $d$ the target quality level, with $s \geq d$, and let $k \in \{s,s-1,\ldots,d\}$ denote the stage index; a lower index implies a more compressed state. Under DRF, as shown in Figure~\ref{fig:Naive_distributed}, each intermediate node performs quality adaptation through pixel-level recompression: receiving compressed data at quality level $s-k+1$, applying decompression operations $D^{s-k+1},g_s^{s-k+1}$, then compression operations $g_a^{s-k},Q^{s-k},E^{s-k}$ to achieve quality level $s-k$. Although it actively utilizes intermediate computational resources, it suffers from redundant pixel-domain conversions and computational inefficiency.

\noindent
\textbf{Fundamental Limitations.}
These frameworks face critical challenges in distributed multi-stage scenarios: PCF underutilizes available computational resources and DRF incurs computational inefficiency through redundant processing cycles. Both approaches treat quality adaptation as separate from the compression process itself, limiting optimization potential for distributed multi-stage compression.

\subsection{Hierarchical Cascade Framework}
\label{subsec:hcf}

To address these fundamental limitations, we propose the Hierarchical Cascade Framework (HCF) that introduces a paradigm shift from adaptation-after-compression to adaptation-during-compression. We assume that the quality levels are ordered with $\lambda_s > \lambda_{s-1} > \cdots > \lambda_d$, where $\lambda_k$ denotes the rate-distortion parameter for quality level $k$, and higher $\lambda$ values correspond to higher compression quality. Instead of treating distributed nodes as passive relay points or applying pixel-domain recompression cycles, HCF establishes direct latent-space transformation pathways that preserve and enhance compression efficiency throughout the distributed pipeline.

\begin{figure*}[t]
    \centering
    \includegraphics[width=\linewidth]{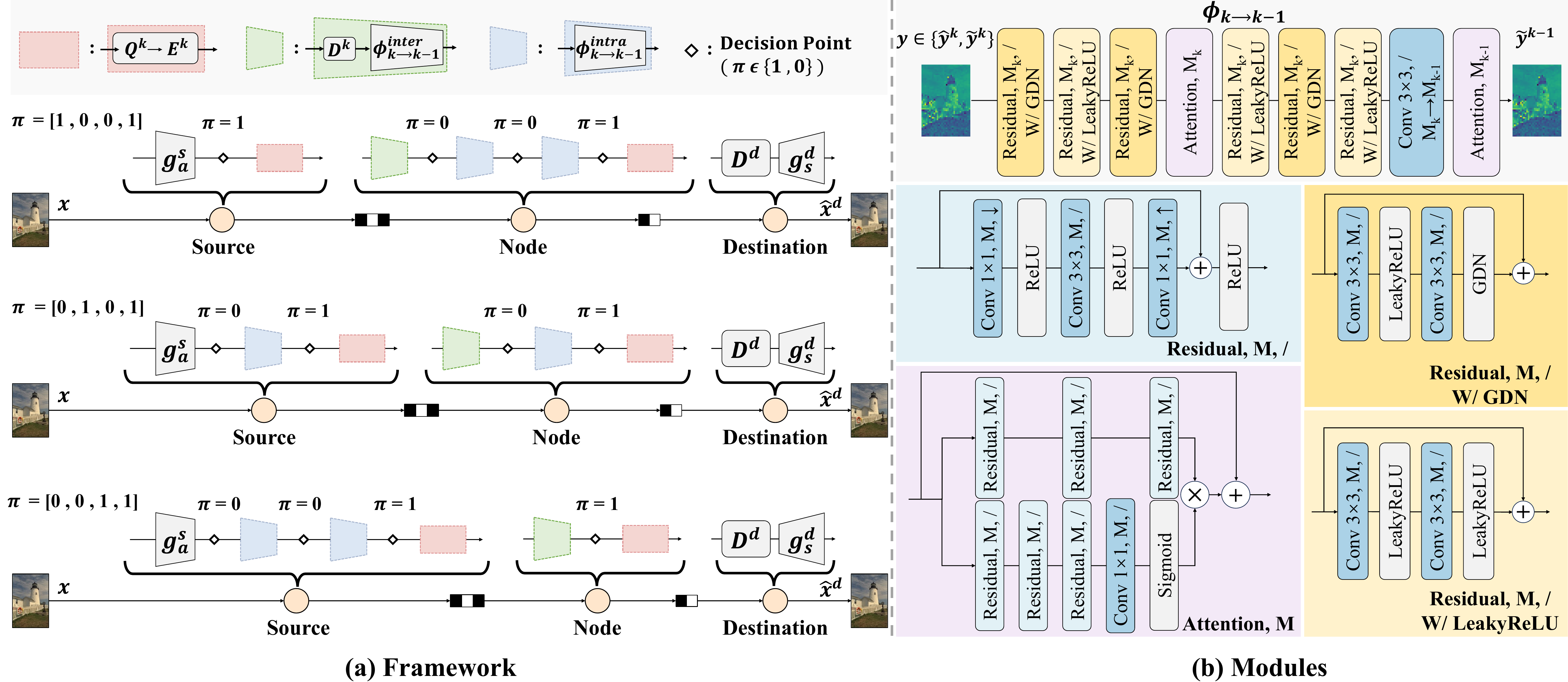}
    \caption{Hierarchical Cascade Framework (HCF). 
    (a) Three $4$-level compression examples in a $2$-stage compression system, each corresponding to a different policy vector 
    $\boldsymbol{\pi} \in \{[1,0,0,1], [0,1,0,1], [0,0,1,1]\}$ that determines the quantization placement in the transformation chain. 
    At each level $k$, the process type follows the definition in Eq.~\eqref{eq:selective_quant}. 
    (b) Architecture of transform modules $\phi_{k \rightarrow k-1}$ designed for cascade processing. 
    $M$ denotes the number of channels in the latent representation. 
    In convolutions, the change of the channel dimension is denoted by $\downarrow$ ($\times \frac{1}{2}$), $\uparrow$ ($\times 2$), and / (unchanged), respectively. 
    The notation $M_k \rightarrow M_{k-1}$ indicates that the convolution maps the latent features from level $k$ to level $k-1$. 
    All convolutions have stride $1$.}

    \label{fig:archi}
\end{figure*}

The framework introduces two key innovations: direct latent-space transformation and selective quantization with policy-driven control. Upon receiving compressed data, an intermediate node may or may not perform compression. For ease of exposition, we focus only on the nodes that perform compression, omitting the others. This design allows a node to compress the received image by multiple levels, depending on available computational resources. Note that if a node compresses the received data by more than two quality levels (e.g., from level 6 directly to level 3), intermediate quantization and entropy encoding operations can be skipped since only the final quantization is transmitted. The analysis transform ($g_a$) and synthesis transform ($g_s$) are applied only at the beginning and end stages, while intermediate nodes apply direct transformations to the latent space. These changes not only simplify the whole operations, but also allow us to selectively conduct the operations at intermediate nodes. Under this framework, we are able to dynamically optimize the overall process of multi-stage compression over the path according to resource availability at the nodes.

We start by defining the process type. From the perspective of the compression operations, we categorize the level-$k$ process into two different types: inter-node process that includes quantization and entropy encoding, and intra-node process that does not.

The \emph{inter-node process} is used when data transmission to the next hop is involved during the level-$k$ process. It consists of the quantization $Q^k$, entropy encoding $E^k$, and entropy decoding $D^k$ operations, followed by transform module $\phi_{k\rightarrow k-1}^{\text{inter}}$. The transmission occurs between the entropy encoding and decoding operations. Given unquantized input $\tilde{y}^{k}$, we present the level-$k$ inter-node process as
\begin{align} \label{eq:inter_node}
\mathcal{T}^{\text{inter}}_{k}(\tilde{y}^{k})
:= \big(\phi^{\text{inter}}_{k \rightarrow k-1} \circ D^{k} \circ E^{k} \circ Q^{k}\big)(\tilde{y}^{k}),
\end{align}
where $\circ$ denotes function composition.
The transform module $\phi^{\text{inter}}_{k \rightarrow k-1}$ is specifically designed for `quantized' input.

The \emph{intra-node process}, in contrast, simply preserves the unquantized input $\tilde{y}^k$ and directly applies a transform module designed for `unquantized' input. We present the level-$k$ intra-node process as
\begin{align} \label{eq:intra_node}
\mathcal{T}^{\text{intra}}_{k}(\tilde{y}^{k}) &:= \phi^{\text{intra}}_{k \rightarrow k-1} (\tilde{y}^{k}).
\end{align}

Note that both the inter- and intra-node processes end with a transform module that shares an identical architecture (see Figure~\ref{fig:archi}(b)) but requires separate training for different input distributions at each level. The novel modules contain residual blocks, attention mechanisms, convolution operations with GDN~\cite{GDN} and LeakyReLU activations, and channel dimension adjustments to handle both quantized and unquantized latent representations. Our primary contribution lies in the development of a systematic framework enabling policy-driven quantization control across multiple compression stages, with the unified architecture facilitating seamless integration across the compression pipeline. Ablation studies (Section~\ref{subsec:ablation}) demonstrate the effectiveness of this design and confirm superior performance in distributed multi-stage scenarios.

The separation of the inter- and intra-node processes allows us to selectively quantize the data at each level $k$, and optimize the compression operations over the network. We define the \emph{policy vector} $\boldsymbol{\pi} = [\pi_s, \pi_{s-1}, \ldots, \pi_{d}]$ where $\pi_k \in \{0,1\}$ indicates the type of level~$k$ process: $\pi_k=1$ for the inter-node process and $\pi_k=0$ for the intra-node process. We fix $\pi_d = 1$ at the last level since the level-$d$ data should be transmitted to the destination and will be decompressed.

Using $\pi_k$, we can present the level-$k$ process in a unified manner as
\begin{equation} \label{eq:selective_quant}
\mathcal{T}_{k}^{\pi_k}(\tilde{y}^{k}) = (1-\pi_k)\mathcal{T}^{\text{intra}}_{k}(\tilde{y}^{k}) + \pi_k\mathcal{T}^{\text{inter}}_{k}(\tilde{y}^{k}).
\end{equation}

\noindent We define the transformation under policy $\boldsymbol{\pi}$ from level $s$ to any intermediate level $k$ as the composition $\mathcal{F}_{s \rightarrow k}^{\boldsymbol{\pi}} = \mathcal{T}_{k+1}^{\pi_{k+1}} \circ \cdots \circ \mathcal{T}_{s}^{\pi_s}$, and the complete transformation from source to destination as
\begin{equation} \label{eq:composite_transform}
    \mathcal{F}_{s \rightarrow d}^{\boldsymbol{\pi}} = \mathcal{T}_{d+1}^{\pi_{d+1}} \circ \mathcal{T}_{d+2}^{\pi_{d+2}} \circ \cdots \circ \mathcal{T}_{s-1}^{\pi_{s-1}} \circ \mathcal{T}_{s}^{\pi_s}.
\end{equation}

\noindent Including the initial process at the source, the last level-$d$ process (without a transform module), and the decompression at the destination, the complete end-to-end cascade process $\mathcal{C}(s,d,\boldsymbol{\pi}): x \mapsto \hat{x}^{d}$ mapping input image $x$ to reconstruction $\hat{x}^{d}$ can be represented as
\begin{equation}
\mathcal{C}(s,d,\boldsymbol{\pi})(x)
= \big(g_s^{d} \circ D^{d} \circ E^{d} \circ Q^{d} \circ \mathcal{F}_{s \rightarrow d}^{\boldsymbol{\pi}} \circ g_a^{s}\big)(x).
\label{eq:cascade_process}
\end{equation}

Figure~\ref{fig:archi} provides a comprehensive visual representation of our HCF framework in a simple network scenario with one intermediate node between the source and destination. 
Both nodes participate in a $4$-level compression process, forming a $2$-stage compression system. 
Accordingly, the policy vector $\boldsymbol{\pi}$ has length $4$ with $\pi_d = 1$. 
In this $2$-stage setting, two transmissions occur along the path, and each transmission must terminate with a quantization operation. 
HCF enables strategic placement of these quantization points within the overall compression pipeline. 
For a $4$-level, $2$-stage configuration, there are three possible quantization placements, $\boldsymbol{\pi} \in \{[1,0,0,1], [0,1,0,1], [0,0,1,1]\}$, since the final level-$d$ process necessarily includes transmission to the destination. 
As the leftmost $1$ in $\boldsymbol{\pi}$ shifts rightward, a larger portion of the compression workload is handled by the source, while the intermediate node processes fewer levels.

\begin{table*}[t]
\centering
\small
\begin{tabular}{@{}cccccccc@{}}
\toprule
\textbf{Quant. Freq. ($n_q$)} & \textbf{Target Qual. ($d$)} & \textbf{Policy Vector ($\boldsymbol{\pi}$)} & \textbf{Bitrate (bpp)} & \textbf{PSNR} $\uparrow$ & \textbf{MS-SSIM} $\uparrow$ & \textbf{$\eta^{\text{PSNR}}$} $\downarrow$ & \textbf{$\eta^{\text{MS-SSIM}}$} $\downarrow$ \\
\midrule
\multirow{2}{*}{2} & \multirow{2}{*}{2} & \textbf{[1,0,0,0,1]} & \textbf{0.1674} & \textbf{30.264} & \textbf{13.129} & \textbf{11.054} & \textbf{13.573} \\
& & [0,0,0,1,1] & 0.1674 & 29.768 & 12.761 & 17.115 & 18.070 \\
\midrule
\multirow{2}{*}{4} & \multirow{2}{*}{1} & \textbf{[1,1,1,0,0,1]} & \textbf{0.0981} & \textbf{28.524} & \textbf{11.249} & \textbf{19.057} & \textbf{22.361} \\
& & [0,0,1,1,1,1] & 0.0951 & 27.929 & 10.819 & 27.419 & 28.010 \\
\bottomrule
\end{tabular}
\caption{Representative quantization policy comparison using MLIC++ with HCF on Kodak dataset. Bold entries show consistently superior performance of $\boldsymbol{\pi}^{\text{edge}}$ strategies ($\uparrow$ higher is better, $\downarrow$ lower is better). Complete analysis across all quantization frequencies, configurations, and additional datasets in Supplementary Material.}
\label{tab:quantization_policy_comparison}
\end{table*}

\subsection{Training Strategy}
\label{sec:training}

Our training strategy consists of two sequential phases: transform module training and fine-tuning. 
We utilize pre-trained single-stage compression model parameters.

\noindent
\textbf{Training of transform modules.} 
We sequentially train transform modules $\phi^{\text{inter}}_{k \rightarrow k-1}$ or $\phi^{\text{intra}}_{k \rightarrow k-1}$ 
for each level pair $(k,k-1)$ from $k=s$ down to $k=d+1$. 
For intra-node modules (unquantized input), we minimize
\begin{equation}
    \mathcal{L}_{\text{intra}}^{k \rightarrow k-1}
    = \big\| \phi^{\text{intra}}_{k \rightarrow k-1}(\tilde{y}^{k}) 
    - g_a^{k-1}(x) \big\|_2^2,
    \label{eq:loss_intra}
\end{equation}
and for inter-node modules (quantized input), we minimize
\begin{equation}
    \mathcal{L}_{\text{inter}}^{k \rightarrow k-1}
    = \big\| \phi^{\text{inter}}_{k \rightarrow k-1}(\hat{y}^{k}) 
    - g_a^{k-1}(x) \big\|_2^2,
    \label{eq:loss_inter}
\end{equation}
where $\tilde{y}^{k} = (\mathcal{F}_{s \rightarrow k}^{\boldsymbol{\pi}} \circ g_a^{s})(x)$
and $\hat{y}^{k} = D^{k}\!\big(E^{k}\!\big(Q^{k}(\tilde{y}^{k})\big)\big)$.
During training, all previously optimized networks are frozen, and only the current transform module is updated. 
Our unified architecture handles both quantized and unquantized inputs via weight switching rather than separate models, minimizing deployment overhead. 
Inspired by knowledge distillation~\cite{hinton2015distilling}, our modules enable adaptive compression across distributed quality levels in multi-stage compression systems.

\noindent
\textbf{Fine-tuning.} 
We perform end-to-end optimization for each level $k \in \{s,s-1,\ldots,d\}$ using the rate–distortion objective:
\begin{equation}
    \mathcal{L}_{\text{RD}}^{k} 
    = \lambda_k \cdot \mathcal{D}(x, \hat{x}^{k}) 
    + \mathcal{R}(\hat{y}^{k}),
    \label{eq:rd_loss}
\end{equation}
where $\tilde{y}^{k} = \big(\mathcal{F}_{s \rightarrow k}^{\boldsymbol{\pi}} \circ g_a^{s}\big)(x)$,
$\hat{y}^{k} = D^{k}\!\big(E^{k}\!\big(Q^{k}(\tilde{y}^{k})\big)\big)$,
and $\hat{x}^{k} = g_s^{k}(\hat{y}^{k})$.
Here, $\mathcal{D}(\cdot,\cdot)$ and $\mathcal{R}(\cdot)$ denote the distortion and rate estimation functions, respectively.
We freeze networks from higher quality levels and all transform modules, optimizing only the synthesis transform $g_s^{k}$ and associated entropy models at level $k$.

\section{Experiments} 
\label{sec:experiments}
\subsection{Experimental Setup}
We validated HCF across five diverse compression architectures—MLIC++~\cite{jiang2023mlicpp}, cheng2020\_attn and cheng2020\_anchor~\cite{cheng2020image}, and mbt2018 and mbt2018\_mean~\cite{minnenbt18}—spanning from hyperprior models to context-attention mechanisms, ensuring generalization across architectural paradigms. We train the models following our two-phase strategy (Section~\ref{sec:training}) using DUTS dataset~\cite{dataset} with standard augmentation. The models are initialized from pre-trained CompressAI~\cite{compressai} weights and official implementations with quality levels $s=6$ (MLIC++, cheng2020\_anchor, cheng2020\_attn) and $s=8$ (mbt2018, mbt2018\_mean). We use Kodak~\cite{kodak_dataset}, CLIC2020-mobile~\cite{clic2020}, and CLIC~\cite{clic2020} datasets for the evaluation.

We systematically compared HCF against four baseline categories: (1) PCF including latest SOTA approaches (Presta~\cite{Presta_2025_WACV}, DeepHQ~\cite{lee2025deephq}, Li~\cite{Li_tcsvt}, Jeon~\cite{2023_CVPR_jeon}, Tian~\cite{Tian_VCIP}, Lee~\cite{Lee_2022_CVPR}) and foundational works~\cite{Toderici,Johnston,Diao}; (2) DRF extending successive compression~\cite{SIC}; (3) SSF as performance bounds; (4) Traditional standards (JPEG~\cite{jpeg}, JPEG2000~\cite{jpeg2000}, BPG~\cite{bpg}, HEVC~\cite{hevc}, WebP~\cite{webp}).

For a fair comparison, DRF and SSF use identical architectures to HCF. 
We report rate–distortion curves and Bjøntegaard Delta metrics~\cite{BJONTEGARD}: 
BD-Rate$_{\text{P}}$/BD-Rate$_{\text{M}}$ (bitrate reduction for PSNR/MS-SSIM~\cite{Wang_msssim}, lower is better) 
and BD-PSNR/BD-MS-SSIM (quality improvement, higher is better), along with computational efficiency metrics including FLOPs, parameters, GPU memory, and execution time reduction relative to the baselines. 
All experiments used NVIDIA RTX A6000 GPUs. The experimental details and hyperparameters are described in Supplementary Material.

\subsection{Policy Analysis}
\label{subsec:policy_analysis}

HCF's flexibility in quantization placement across multiple compression stages naturally raises questions about optimal policy selection. To address this systematically, we establish a mathematical framework for policy representation. For an $n_q$-quantization compression system (where $n_q$ denotes the number of quantization operations) with quality levels from $s$ to $d$, we define the policy space as all feasible quantization placement patterns with exactly $n_q$ quantization operations, which is formally given by
\begin{equation}
  \scalebox{0.96}{$\displaystyle \Pi(s,d;n_q) := \left\{\boldsymbol{\pi} \in \{0,1\}^{s-d+1} \mid \sum_{k=d}^{s}\boldsymbol{\pi}_k=n_q, \, \pi_d = 1\right\}.$}\nonumber
\end{equation}
Given the significant variation in rate-distortion performance across different policies within the same $n_q$ constraint, we evaluated policy effectiveness using established quality metrics (e.g., PSNR, MS-SSIM). To gain deeper insights into the fundamental principles governing optimal quantization placement, we developed a Rate-Quality Sensitivity Index (RQSI) for analyzing quantization efficiency.

\noindent
\textbf{Rate-Quality Sensitivity Index (RQSI).} We quantified policy efficiency using reference vectors: $\boldsymbol{\pi}_* = [0,\ldots,0,1] \in \Pi(s,d;1)$ (minimal quantization, only at the final level) and $\boldsymbol{\pi}^* = [1,\ldots,1,1] \in \Pi(s,d;s-d+1)$ (maximal quantization, at every level). Let $\mathcal{M}$ denote the quality metric (i.e., PSNR or MS-SSIM), and let $\hat{x}^{d+1}_{\boldsymbol{\pi}}$, $\hat{x}^{d}_{\boldsymbol{\pi}}$ and $\hat{y}^{d+1}_{\boldsymbol{\pi}}$, $\hat{y}^{d}_{\boldsymbol{\pi}}$ denote reconstructed images and quantized latent representations at quality levels $d+1$ and $d$ under policy $\boldsymbol{\pi}$. We define the RQSI of policy $\boldsymbol{\pi}$ as 
\begin{equation}
    \eta^{\mathcal{M}}(\boldsymbol{\pi}) = \textstyle \frac{1}{2}\left( RQS(\boldsymbol{\pi},\boldsymbol{\pi}_*) + RQS(\boldsymbol{\pi},\boldsymbol{\pi}^*) \right),
\end{equation}
where
\begin{equation}
RQS(\boldsymbol{\pi}_1,\boldsymbol{\pi}_2)=\frac{\big|\mathcal{M}(x,\hat{x}^{d+1}_{\boldsymbol{\pi}_2}) - \mathcal{M}(x,\hat{x}^d_{\boldsymbol{\pi}_1})\big|}{\max\big(\big| \mathcal{R}(\hat{y}^{d+1}_{\boldsymbol{\pi}_2}) - \mathcal{R}(\hat{y}^d_{\boldsymbol{\pi}_1})\big|, \varepsilon \big)},
\end{equation}
\noindent 
and $\varepsilon > 0$ is a small constant ensuring numerical stability (e.g., $\varepsilon = 10^{-6}$). The RQSI metric $\eta^{\mathcal{M}}$ provides a measure to compare quantization placement strategies, with lower values indicating superior rate-quality efficiency. By benchmarking against $\boldsymbol{\pi}_*$ and $\boldsymbol{\pi}^*$, RQSI isolates the impact of quantization placement within the cascade, independent of the compression level. This identifies policies that preserve critical information for downstream transformations.

We conducted comprehensive experiments using MLIC++ on the Kodak dataset across different quantization frequencies. Table~\ref{tab:quantization_policy_comparison} presents representative comparisons between two different policies, with bold entries indicating superior performance. For $n_q=2$ and target quality $d=2$, the policy $[1,0,0,0,1]$ achieved 30.264 dB PSNR compared to 29.768 dB for the alternative policy $[0,0,0,1,1]$ at identical bitrates (0.50 dB improvement). For the more complex scenario with $n_q=4$ and $d=1$, the policy $[1,1,1,0,0,1]$ achieves 28.524 dB PSNR versus 27.929 dB for $[0,0,1,1,1,1]$ (0.60 dB gain) while using only marginally higher bitrate (0.0981 vs 0.0951 bpp). Comprehensive results across all quantization frequencies ($n_q$) and target quality levels ($d$) are in Supplementary Material.

In these examples, we can observe that early-quantization strategies--those that `1's appear at early stage, i.e., $[1,0,0,0,1]$ and $[1,1,1,0,0,1]$--outperform their counterparts. This, indeed, has been observed throughout our experiments. The RQSI metric further clarifies the observation by showing that the edge policy achieves a lower RQSI value: $\eta^{\text{PSNR}}([1,0,0,0,1])=11.054 < \eta^{\text{PSNR}}([0,0,0,1,1])=17.115$ and $\eta^{\text{PSNR}}([1,1,1,0,0,1])=19.057 < \eta^{\text{PSNR}}([0,0,1,1,1,1])=27.419$.

We formally define such early-quantization policy as \emph{edge policy} $\pi^{\text{edge}}$ as 
\begin{equation}
\boldsymbol{\pi}^{\text{edge}} = [1^{(n_q-1)}, 0^{(s-d+1-n_q)}, 1],
\end{equation}
where $1^k$ and $0^k$ denote $k$ consecutive ones and zeros (here $k=n_q-1$ and $k=s-d+1-n_q$ respectively). This strategically places quantization at cascade edges: $(n_q-1)$ operations at front stages plus one mandatory final operation. Evaluation across all quantization frequencies and configurations (detailed in Supplementary Material) consistently validated this edge quantization principle.

\begin{figure}[t]
    \centering
    \includegraphics[width=\linewidth]{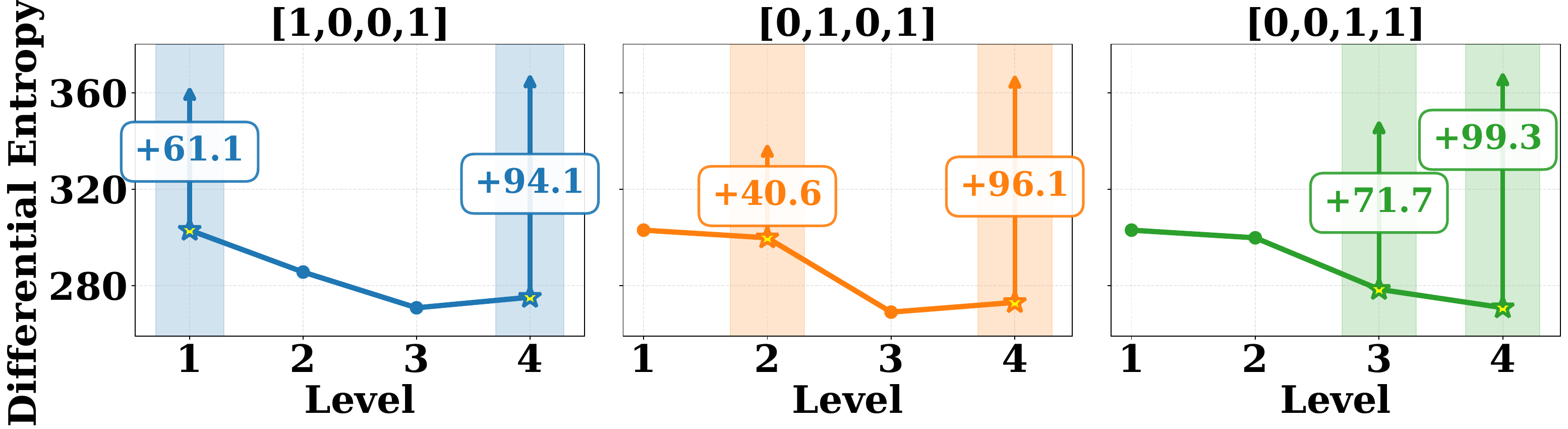}
    \caption{Differential entropy evolution across compression levels for HCF's three quantization policies using MLIC++ on Kodak in a 2-stage, 4-level system. Stars indicate quantization points; $\uparrow$ show entropy increases after quantization.}
    \label{fig:entropy_evolution}
\end{figure}

\begin{figure}[t]
    \centering
    \includegraphics[width=\linewidth]{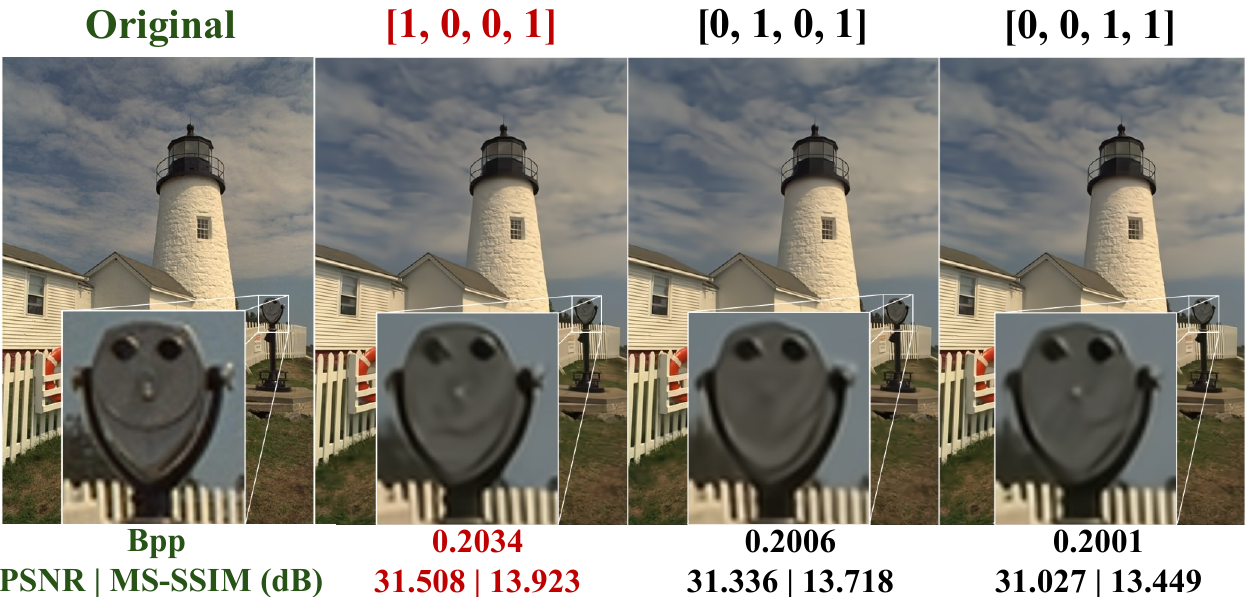}
    \caption{Visual quality comparison of HCF policies on Kodak dataset using MLIC++. $\boldsymbol{\pi}^{\text{edge}}=[1,0,0,1]$ preserves more details compared to alternatives at equivalent bitrates.}
    \label{fig:image_vis1}
\end{figure}

To understand the underlying mechanism driving this performance difference, we analyzed differential entropy evolution across compression levels using the Kozachenko-Leonenko estimator~\cite{KozLeo87}. Figure~\ref{fig:entropy_evolution} shows differential entropy evolution across four compression levels for three quantization policies. While all policies converge to similar entropy values before the final quantization operation, they exhibit markedly different entropy increases after quantization: 94.1, 96.1, and 99.3 bits for policies [1,0,0,1], [0,1,0,1], and [0,0,1,1], respectively. Policy $\boldsymbol{\pi}^{\text{edge}}=[1,0,0,1]$ achieved the smallest increment, demonstrating that early quantization injection allows subsequent transform modules to more effectively decorrelate and suppress redundant information, thereby minimizing the irreducible uncertainty that must be encoded. This theoretical advantage translated into visually perceptible improvements, as demonstrated in Figure~\ref{fig:image_vis1}, where $\boldsymbol{\pi}^{\text{edge}}$ preserves significantly more structural and textural details compared to alternative policies at equivalent bitrates. Comprehensive entropy analysis across different quality configurations (detailed in Supplementary Material) confirms that this pattern consistently holds across diverse scenarios.

\subsection{Performance Comparison}
\label{subsec:performance_comparison}

Based on the superior $\boldsymbol{\pi}^{\text{edge}}$ policy identified through systematic analysis across multiple quantization frequencies, we evaluated HCF's overall performance against established baselines. We focus on the widely-applicable $n_q=2$ configuration, which represents the most common two-stage compression scenario in practical distributed systems (e.g., source $\rightarrow$ base station $\rightarrow$ destination), while maintaining computational tractability. This configuration consistently demonstrates the edge quantization principle across all tested architectures, with similar performance patterns observed for higher quantization frequencies ($n_q>2$) as detailed in Supplementary Material.

\begin{table}[t]
\centering
\small
\begin{tabular}{ccccc}
\toprule
\multirow{3}{*}{\textbf{Model}} & 
\multicolumn{2}{c}{\textbf{Kodak}} & 
\multicolumn{2}{c}{\textbf{CLIC}} \\
\cmidrule(lr){2-3}\cmidrule(lr){4-5}
& \textbf{BD-Rate$_{\text{P}}$} & \textbf{BD-PSNR} & \textbf{BD-Rate$_{\text{P}}$} & \textbf{BD-PSNR} \\
& \textbf{(\%)} $\downarrow$ & \textbf{(dB)} $\uparrow$ & \textbf{(\%)} $\downarrow$ & \textbf{(dB)} $\uparrow$ \\
\midrule
Presta & +12.64 & -0.46 & +9.48 & -0.28 \\
Jeon & +13.40 & -0.51 & +8.62 & -0.27 \\
Lee & +43.84 & -1.76 & +26.39 & -0.80 \\
\midrule
DRF & +4.87 & -0.22 & +5.56 & -0.23 \\
\bottomrule
\end{tabular}
\caption{BD-metric evaluation relative to HCF with $\boldsymbol{\pi}^{\text{edge}}$ policy using MLIC++. DRF is the MLIC++ DRF variant. Others are PCF variants. $\uparrow$ higher is better, $\downarrow$ lower is better.}
\label{tab:method_comparison}
\end{table}

\begin{figure}[t]
    \centering
    \includegraphics[width=\linewidth]{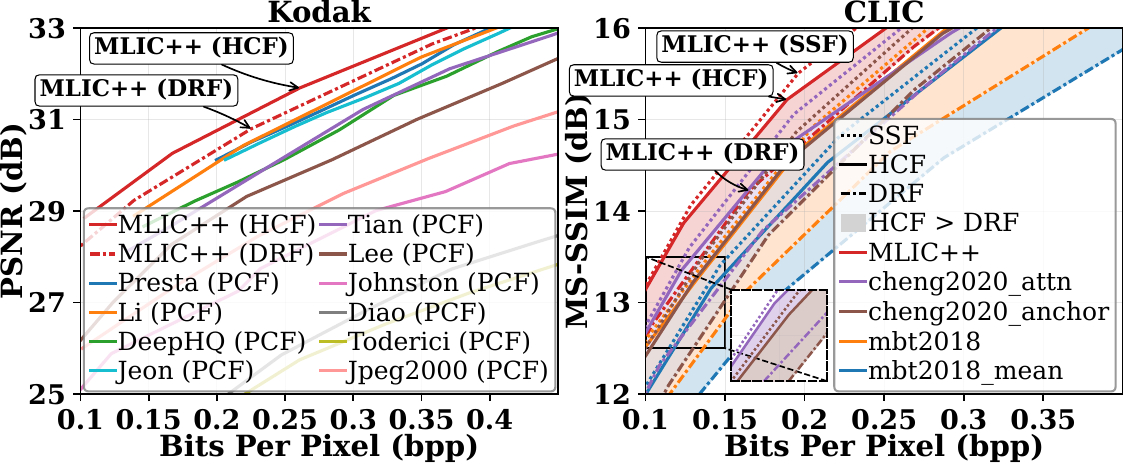}
    \caption{Rate-distortion comparison using PSNR on Kodak dataset and MS-SSIM on CLIC dataset. HCF with $\boldsymbol{\pi}^{\text{edge}}$ policy outperforms PCF and DRF baselines while approaching SSF performance. Shaded areas indicate HCF's performance advantage over DRF. Upper-left is better.}
    \label{fig:RD_curve}
\end{figure}

\begin{table*}[t]
\centering
\small
\begin{tabular}{ccccccc}
\toprule
\multirow{2}{*}{\textbf{Model}} & \multicolumn{5}{c}{\textit{FLOPs Reduction / Parameters Reduction / GPU Memory Reduction / Execution Time Reduction (\%) $\uparrow$}} \\
\cmidrule{2-6}
& \textbf{Node$_1$} & \textbf{Node$_2$} & \textbf{Node$_3$} & \textbf{Node$_4$} & \textbf{Node$_5$}\\
\midrule
cheng2020\_attn & 97.8/68.7/95.5/90.0 & 97.8/68.7/95.5/90.0 & 97.4/65.8/95.0/87.6 & 97.8/68.7/96.5/83.5 & 97.8/68.7/96.5/83.5\\
cheng2020\_anchor & 97.5/62.7/95.4/87.1 & 97.5/62.7/95.4/87.1 & 97.1/60.2/94.8/83.9 & 97.5/62.7/96.5/77.9 & 97.5/62.7/96.5/77.9\\
\bottomrule
\end{tabular}
\caption{Computational efficiency comparison: HCF vs. DRF across five processing nodes. Values show HCF's percentage reduction in FLOPs/Parameters/GPU Memory/Execution Time relative to DRF for transform operations. $\uparrow$ higher is better.}
\label{tab:computational_efficiency}
\end{table*}

Figure~\ref{fig:RD_curve} presents comprehensive rate-distortion comparisons across five compression architectures: SSF (centralized baseline), DRF (distributed recompression baseline), PCF variants (bitstream truncation), and HCF using $\boldsymbol{\pi}^{\text{edge}}$. HCF demonstrates superior performance over distributed baselines while approaching centralized SSF performance, with consistent cross-architecture improvements evidencing our approach's generalizability. Table~\ref{tab:method_comparison} quantifies these improvements using MLIC++ architecture under fair comparison conditions—relative to HCF with MLIC++ using $\boldsymbol{\pi}^{\text{edge}}$ policy, PCF methods exhibited BD-Rate$_{\text{P}}$ increases of 12.64\% (Presta), 13.40\% (Jeon), and 43.84\% (Lee) on Kodak with similar CLIC trends, while DRF shows 4.87\% (Kodak) and 5.56\% (CLIC) increases, demonstrating direct latent-space transformations' effectiveness over bitstream truncation and pixel-domain recompression. Beyond rate-distortion gains, HCF achieved substantial computational efficiency through eliminating redundant decode-encode operations: Table~\ref{tab:computational_efficiency} shows overall reductions of 97.1--97.8\% FLOPs, 94.8--96.5\% GPU memory, and 77.9--90.0\% execution time across evaluated models. Additional results across datasets and architectures are in Supplementary Material.

\subsection{Cross-Quality Adaptation and Error Control}
\label{subsec:adaptation_error_control}
HCF enables retraining-free adaptation across quality levels through learned latent-space transformations. These transformations provide error control across compression trajectories. Our framework supports multiple compression paths using a single model, eliminating the need for separate training for each quality configuration. Table~\ref{tab:cross_quality} demonstrates this using cheng2020\_attn on CLIC2020-mobile, where shorter compression paths (e.g., $5 \rightarrow 1$) achieve better BD-Rate$_{\text{P}}$ reductions by 7.13--10.87\% over longer paths (e.g., $6 \rightarrow 1$) to the same target quality. This validated HCF's effective error accumulation control—longer spans introduce more cumulative distortion, while our learned transformations mitigate degradation without retraining, with additional results across architectures provided in Supplementary Material.

\begin{table}[t]
\centering
\small
\begin{tabular}{@{}cccc@{}}
\toprule
\multirow{2}{*}{\textbf{Model}} & \textbf{Compression Path} & \textbf{BD-Rate$_{\text{P}}$} & \textbf{BD-Rate$_{\text{M}}$} \\
& \textbf{Comparison} & \textbf{(\%)} $\downarrow$ & \textbf{(\%)} $\downarrow$ \\
\midrule
\multirow{3}{*}{cheng2020\_attn}
& $5 \rightarrow 1$ vs. $6 \rightarrow 1$ & -7.13 & -7.29 \\
& $4 \rightarrow 1$ vs. $5 \rightarrow 1$ & -7.36 & -7.02 \\
& $3 \rightarrow 1$ vs. $4 \rightarrow 1$ & -10.87 & -7.80 \\
\bottomrule
\end{tabular}
\caption{HCF cross-quality adaptation without retraining on CLIC2020-mobile dataset. BD-Rate reductions validate controlled error accumulation. $\downarrow$ lower is better.}
\label{tab:cross_quality}
\end{table}

\subsection{Ablation Studies}
\label{subsec:ablation}

\begin{figure}[t]
\centering
\includegraphics[width=1\linewidth]{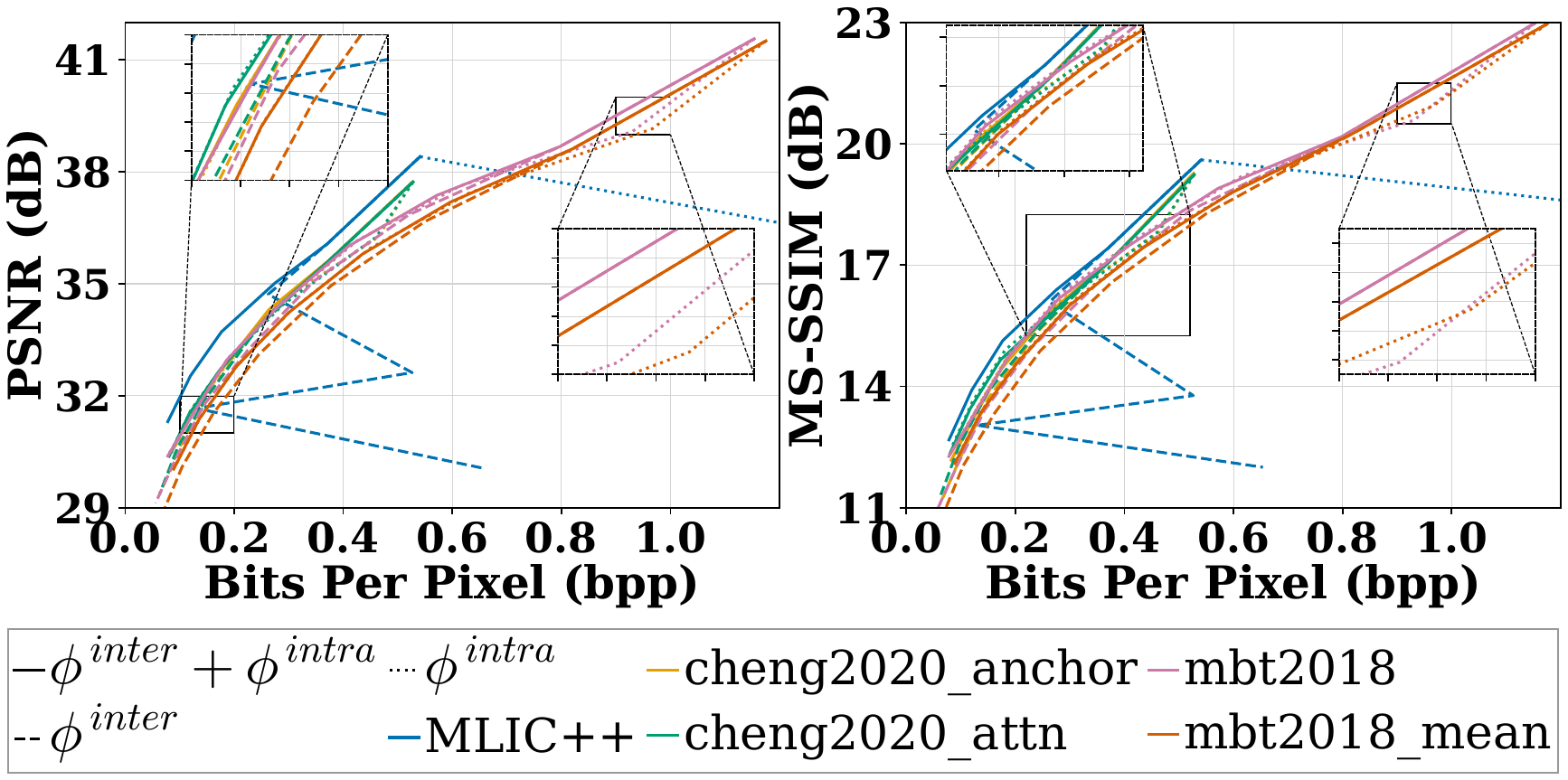}
\caption{Ablation study comparing complete HCF ($\phi^{\text{inter}} + \phi^{\text{intra}}$) with $\phi^{\text{inter}}$-only and $\phi^{\text{intra}}$-only variants on CLIC2020-mobile dataset under $\boldsymbol{\pi}^{\text{edge}}$ policy. Both modules are essential for superior performance. Upper-left is better.}
\label{fig:ablation_clic2020_mobile}
\end{figure}

To validate our design hypothesis that both transform modules are essential and complementary, we compared three HCF configurations: complete HCF ($\phi^{\text{inter}} + \phi^{\text{intra}}$), $\phi^{\text{inter}}$-only, and $\phi^{\text{intra}}$-only. All use $\boldsymbol{\pi}^{\text{edge}}$ policy. Figure~\ref{fig:ablation_clic2020_mobile} shows that removing either module leads to performance degradation: $\phi^{\text{inter}}$-only suffered at lower bitrates due to inadequate information preservation, while $\phi^{\text{intra}}$-only fails at higher bitrates due to poor quantization artifact handling. These results confirmed that both modules address distinct challenges and their combination is essential for superior performance, with additional ablation studies in Supplementary.

\section{Conclusion}
\label{sec:conclusion}
We propose Hierarchical Cascade Framework (HCF), a novel framework enabling direct latent-space transformations for distributed multi-stage image compression. The framework introduces policy-driven quantization control with optimal strategies ($\boldsymbol{\pi}^{\text{edge}}$) that outperform SOTA methods by up to 12.64\% BD-Rate in PSNR and improve computational efficiency (e.g., FLOPs reduction up to 97.8\%). HCF delivers superior rate-distortion performance while enabling retraining-free cross-quality adaptation. This work contributes to establishing a new paradigm for learned compression across transmission stages in intelligent multimedia systems. Future extensions will target video compression scenarios and intelligent agents for adaptive policy control.

\section*{Acknowledgments}
This work is supported in part by NRF grant funded by the Korea government (MSIT) (RS-2022-NR070834, 33\%), by IITP grant funded by the Korea government (MSIT) (RS-2024-00405128, 33\%), and by the ICT Creative Consilience program through IITP grant funded by the Korea government (MSIT) (IITP-2025-RS-2020-II201819, 33\%).
\bibliography{aaai2026}

\end{document}